\documentclass[sigconf]{acmart}

\usepackage{booktabs} 
\usepackage{mathtools}
\usepackage{float}
\usepackage{graphicx}
\usepackage{subcaption}
\usepackage{multirow}
\usepackage{rotating}
\usepackage{array}
\usepackage{gensymb}
\usepackage{pdfpages}
\usepackage{multirow}
\usepackage{makecell}
\usepackage{wrapfig}
\usepackage{algorithm}
\usepackage[noend]{algpseudocode}
\usepackage{tikz}
\usetikzlibrary[topaths]
\usetikzlibrary{arrows,decorations.markings,arrows.meta}
\usetikzlibrary{positioning}
\tikzset{
    position/.style args={#1:#2 from #3}{
        at=(#3.#1), anchor=#1+180, shift=(#1:#2)
    }
}

\newcommand{\mc}[1]{\mathcal{#1}}
\newcount\mycount
\newcount\mynode

\newcommand{\mycomment}[1]{}

\makeatletter
\newsavebox\myboxA
\newsavebox\myboxB
\newlength\mylenA

\def\vec#1{\mathchoice{\mbox{\boldmath$\displaystyle#1$}}
{\mbox{\boldmath$\textstyle#1$}}
{\mbox{\boldmath$\scriptstyle#1$}}
{\mbox{\boldmath$\scriptscriptstyle#1$}}}
\makeatother

\makeatletter
\newcommand*{\bigominus}{\DOTSB\bigominus@\slimits@}

\newcommand{\bigominus@}{\mathop{\mathpalette\bigominus@@\relax}}

\newcommand{\bigominus@@}[2]{%
  \vcenter{\hbox{%
    \sbox\z@{$\m@th#1\bigoplus$}%
    \resizebox{\wd\z@}{!}{$\m@th#1\bm{\ominus}$}%
  }}%
}
\makeatother

\algblock{ParFor}{EndParFor}
\algnewcommand\algorithmicparfor{\textbf{parfor}}
\algnewcommand\algorithmicpardo{\textbf{do}}
\algnewcommand\algorithmicendparfor{\textbf{end\ parfor}}
\algrenewtext{ParFor}[1]{\algorithmicparfor\ #1\ \algorithmicpardo}
\algrenewtext{EndParFor}{\algorithmicendparfor}

\makeatletter
\ifthenelse{\equal{\ALG@noend}{t}}%
  {\algtext*{EndParFor}}
  {}%
\makeatother

\usepackage{xcolor}
\definecolor{LightGray}{gray}{0.92}
\usepackage[frozencache]{minted}
\setminted[python]{fontsize=\footnotesize,
mathescape,resetmargins,breaklines,
samepage,xleftmargin=0px
}

\copyrightyear{2023}
\acmYear{2023}
\setcopyright{rightsretained}
\acmConference[GECCO '23 Companion]{Genetic and Evolutionary Computation Conference Companion}{July 15--19, 2023}{Lisbon, Portugal}
\acmBooktitle{Genetic and Evolutionary Computation Conference Companion (GECCO '23 Companion), July 15--19, 2023, Lisbon, Portugal}\acmDOI{10.1145/3583133.3596361}
\acmISBN{979-8-4007-0120-7/23/07}

\begin{document}

\title{A Joint Python/C++ Library for Efficient yet Accessible Black-Box and Gray-Box Optimization with GOMEA}

\author{Anton Bouter}
\affiliation{%
  \institution{Centrum Wiskunde \& Informatica}
  \city{Amsterdam}
  \country{The Netherlands}
}
\email{Anton.Bouter@cwi.nl}

\author{Peter A.N. Bosman}
\affiliation{%
  \institution{Centrum Wiskunde \& Informatica}
  \city{Amsterdam}
  \country{The Netherlands}
}
\email{Peter.Bosman@cwi.nl}

\begin{abstract}
Exploiting knowledge about the structure of a problem can greatly benefit the efficiency and scalability of an Evolutionary Algorithm (EA).
Model-Based EAs (MBEAs) are capable of doing this by explicitly modeling the problem structure.
The Gene-pool Optimal Mixing Evolutionary Algorithm (GOMEA) is among the state-of-the-art of MBEAs due to its use of a linkage model and the optimal mixing variation operator. 
Especially in a Gray-Box Optimization (GBO) setting that allows for partial evaluations, i.e., the relatively efficient evaluation of a partial modification of a solution, GOMEA is known to excel.
Such GBO settings are known to exist in various real-world applications to which GOMEA has successfully been applied.
In this work, we introduce the \texttt{GOMEA} library, making existing GOMEA code in C++ accessible through Python, which serves as a centralized way of maintaining and distributing code of GOMEA for various optimization domains.
Moreover, it allows for the straightforward definition of BBO as well as GBO fitness functions within Python, which are called from the C++ optimization code for each required (partial) evaluation.
We describe the structure of the \texttt{GOMEA} library and how it can be used, and we show its performance in both GBO and Black-Box Optimization (BBO).
\end{abstract}



\maketitle

\section{Introduction}
For many difficult optimization problems it is known that exploiting the problem structure is essential for an Evolutionary Algorithm (EA) to achieve good performance and scalability.
A class of EAs that is known to be capable of capturing and exploiting problem structure is that of Model-Based Evolutionary Algorithms (MBEAs) \cite{cheng2018model,bull1999model}.
An algorithm that is among the state of the art among MBEAs, and in the field of Evolutionary Computation (EC) in general, is the Gene-pool Optimal Mixing Evolutionary Algorithm (GOMEA) \cite{thierens2011optimal,dushatskiy2021parameterless,bouter2021achieving}, which explicitly models the problem structure of a problem using a linkage model, and exploits this model when applying variation using the Gene-pool Optimal Mixing (GOM) variation operator.

In general, GOMEA is capable of learning the structure of the optimization problem during optimization.
However, for many problems it is known \textit{a priori} what the rough problem structure is.
In such cases, it has been shown that this knowledge can greatly benefit the efficiency of EAs, both for benchmarks problems \cite{tintos2015partition,chen2018tunneling,bouter2021achieving} as well as various real-world applications within the medical field \cite{bouter2019gpu}, engineering \cite{deb2016breaking}, and vehicle routing \cite{whitley2009tunneling}.
Such a setting where a limited amount of domain knowledge is available and can be used by the optimization algorithm is called a Gray-Box Optimization (GBO) setting, in contrast to a Black-Box Optimization (BBO) setting where no domain knowledge is available.
In particular, we consider the GBO setting where it is known how the fitness function is constructed from a number of subfunctions, and partial evaluations are possible.
Such partial evaluations are relatively efficient function evaluations that are used to update the fitness of a solution after a (small) subset of its variables have been modified.
Especially for the GBO setting that allows for partial evaluations, GOMEA has shown to achieve excellent performance and scalability \cite{bouter2021achieving}.

In this work, we introduce the \texttt{GOMEA} library, which is a Python library that wraps optimization code of GOMEA written in C++.
This C++ code is based on the original code provided by the authors of the most recent relevant publications of GOMEA within the respective domains \cite{dushatskiy2021parameterless,bouter2021achieving}.
The \texttt{GOMEA} library serves the purpose of being a centralized way of distributing and maintaining the code of GOMEA, and it makes it easier to install and run GOMEA on user-specific problems.
This is firstly the case because this library can be used within Python, one of the currently most commonly used programming languages, and secondly because it can be straightforwardly installed through the Python package installer \texttt{pip}, i.e., by running \texttt{pip install gomea}.

Furthermore, the \texttt{GOMEA} library supports the user to implement both BBO and GBO problems within Python, which are called from the C++ optimization code whenever a (partial) evaluation is required.
The implementation of a GBO function within the \texttt{GOMEA} library is more straightforward than before, as it only requires the user to define the input and output of each subfunction, rather than knowing the inner workings of the GOMEA code.
Further customization is possible for the implementation of more advanced GBO functions, but can be omitted for relatively simple GBO problems.
For the integration between Python and C++ code, Cython \cite{behnel2011cython} is used, which is also used in many commonly used Python packages, including NumPy \cite{harris2020array}, SciPy \cite{2020SciPy-NMeth}, and Pandas \cite{mckinney-proc-scipy-2010}.

Note that this paper describes the state of the \texttt{GOMEA} library version 1.0, and details are subject to change.
The source code of the \texttt{GOMEA} library is publicly available on GitHub\footnote{\url{https://github.com/abouter/gomea/}}.

In the remainder of this paper, essential background work, including that required to understand and implement a GBO problem within the \texttt{GOMEA} library, is discussed in Section \ref{sec:background}.
In Section \ref{sec:implementation}, the architecture and its implementation are described.
The main features of the \texttt{GOMEA} library, and how they are used, are then described in Section \ref{sec:features}.
To show the general performance of the \texttt{GOMEA} library in both a BBO and a GBO setting, experimental results are shown in Section \ref{sec:experiments}.
Finally, future work is outlined and conclusions are drawn in Sections \ref{sec:discussion} and \ref{sec:conclusion}, respectively.

\section{Background}
\label{sec:background}
We consider optimization problems where the variables to optimize are denoted $\vec{X} = [X_0,X_1,\dots,X_{\ell-1}]$.
The problem variables are indexed through $\mc{I} = [0,1,\dots,\ell-1]$, and a realization of the problem variables is denoted $\vec{x} = \{x_0,x_1,\dots,x_{\ell-1}\}$.
Note that zero-based indexing is used here to be consistent with implementation details to be discussed later.

We consider both the domains of discrete optimization ($\vec{x} \in \mathbb{Z}^\ell$) and real-valued optimization ($\vec{x} \in \mathbb{R}^\ell$).
The objective, or fitness, function $f$ is either subject to minimization in real-valued optimization, or maximization in discrete optimization, corresponding to the respective conventions in these fields.

In this work, we only consider the GBO setting that allows for partial evaluations.
Such a setting is formally defined in Section \ref{subsec:gbo}.

\subsection{Gray-Box Optimization}
\label{subsec:gbo}
As previously defined in \cite{bouter2018large}, an objective function in a GBO setting can be written as:
\begin{align}
\label{eq:gbo}
f(\vec{x}) &= g\left(f_0(\vec{x}_{\varmathbb{I}_0}) \oplus f_1(\vec{x}_{\varmathbb{I}_1}) \oplus \dots \oplus f_{q-1}(\vec{x}_{\varmathbb{I}_{q-1}}) \right),\\
&= g\left( \bigoplus_{i=0}^{q-1} \left( f_i(\vec{x}_{\varmathbb{I}_i}) \right) \right)
\end{align}
where each $f_i(\vec{x}_{\varmathbb{I}_i})$ a subfunction that depends on a subset of variables of $\vec{x}$, namely those for which the index $u$ is included in $\varmathbb{I}_i \subseteq \mc{I}$.
The set $\varmathbb{I} = \{\varmathbb{I}_0, \varmathbb{I}_1, \dots, \varmathbb{I}_{q-1}\}$ defines the complete set of dependencies of each subfunction.
Furthermore, $\oplus$ is a commutative binary operator with a known inverse $\ominus$ (e.g., addition or multiplication), and $g : \mathbb{R} \rightarrow \mathbb{R}$ is any (potentially non-linear) function.

A partial evaluation is a relatively efficient evaluation following the modification of a (small) number of variables of a solution.
Consider a solution $\vec{x}$ in a certain state $g$, denoted $\vec{x}^g$, for which the fitness $f(\vec{x}^{g})$ is known.
After the modification of some variable $x_u$, we denote that this solution is in the state $\vec{x}^{g+1}$.
Performing the partial evaluation following the modification of these variables first requires finding all $\varmathbb{I}_j$ that contain $u$, as these indicate the subfunctions dependent on variables $x_u$ or $x_v$.
For each of these dependent subfunctions, the value given state $\vec{x}^{g}$ should be subtracted from the fitness, and the value given state $\vec{x}^{g+1}$ should be added.
As such, the fitness of $\vec{x}^{g+1}$ is computed as follows:
\begin{align}
\label{eq:parteval}
f\left(\vec{x}^{g+1}\right) &= g\left( \bigoplus_{i=0}^{q-1} f_i(\vec{x}^{g}_{\varmathbb{I}_i}) \oplus \bigoplus_{\varmathbb{I}_i \ni u} f_i(\vec{x}^{g+1}_{\varmathbb{I}_i}) \ominus \bigoplus_{\varmathbb{I}_i \ni u} f_i(\vec{x}^{g}_{\varmathbb{I}_i}) \right),
\end{align}
where $\varmathbb{I}_i \ni u$ is shorthand for the set $\{\varmathbb{I}_i \in \varmathbb{I}~\vert~u \in \varmathbb{I}_i\}$.
However, because all subfunctions for state $\vec{x}^{g}$ have previously already been computed, this computation can be substantially accelerated by storing their sum in memory.
Here, this sum is named a fitness buffer and denoted $\beta$.
This fitness buffer is computed after initialization for each solution in the population, and continuously updated after any modification to the population.

Consider the fitness buffer $\beta$ of solution $\vec{x}$.
For the initial state of the solution, i.e., $\vec{x}^0$, the initial state of the buffer, i.e., $\beta^0$, is computed as the sum of all subfunctions as follows:
\begin{align}
\beta^0 &= \bigoplus_{i=0}^{q-1} f_i(\vec{x}^{0}_{\varmathbb{I}_i}).
\end{align}
Then, following a modification to variable $x_u$, all current values of dependent subfunctions are subtracted from the fitness buffer, and new values are computed and added to the fitness buffer:
\begin{align}
\label{eq:bufferupdate}
\beta^{g+1} &= \beta^{g} \ominus \bigoplus_{\varmathbb{I}_i \ni u} f_i(\vec{x}^{g}_{\varmathbb{I}_i}) \oplus \bigoplus_{\varmathbb{I}_i \ni u} f_i(\vec{x}^{g+1}_{\varmathbb{I}_i})
\end{align}
The update of the fitness buffer is similar when more than one variable is updated.
In this case, the update considers the union of subfunctions dependent on any of the modified variables.

Note the similarities between Equations \ref{eq:parteval} and \ref{eq:bufferupdate}.
Therefore, following an update to the fitness buffer $\beta^{g+1}$, the fitness of solution $\vec{x}^{g+1}$ can be computed in constant time (with respect to the problem size $\ell$) as follows:
\begin{align}
f(\vec{x}^{g+1}) &= g( \beta^{g+1} )
\end{align}

A GBO function is not required to use only one fitness buffer, but can potentially use an arbitrary number of fitness buffers.
As such, it is possible to define GBO functions similar to any of the following signatures:
\begin{align}
\label{eq:gfunc_examples}
f_{\text{Example1}}(\vec{x}) &= g( \beta_0, \beta_1, \beta_2, \dots ) \\
f_{\text{Example2}}(\vec{x}) &= g_0( \beta_0 ) + g_1( \beta_1 ) + g_2( \beta_2 ) + \dots \\
f_{\text{Example3}}(\vec{x}) &= h( g_0( \beta_0, \beta_1 ) + g_1( \beta_2 ) )
\end{align}
Note that none of these functions directly use problem variables, but only fitness buffers, because their complexity should not scale with the number of problem variables in order to maintain the relative efficiency of the GBO setting.

Due to saving the fitness buffer(s) in memory, the complexity of a partial evaluation scales with the number of dependent subfunctions.
Therefore, a partial evaluation that requires the computation of $k$ subfunction is considered to be the fraction $k/q$ of an evaluation.
It is assumed that all subfunctions have approximately the same computational complexity.

\subsection{GOMEA}
The Gene-pool Optimal Mixing Evolutionary Algorithm (GOMEA) \cite{thierens2011optimal} is a Model-Based Evolutionary Algorithm (MBEA) that excels at using domain knowledge of the optimization problem to improve the performance and scalability of the optimization.
This domain knowledge can be learned based on the population during optimization, or, when the problem allows it, be supplied \textit{a priori}.

The dependency structure of a problem is modeled with what is called a linkage model, described in more detail in Section \ref{subsec:linkagemodels}.
This linkage model is then used to guide the Gene-pool Optimal Mixing (GOM) variation operator, which applies variation to only a small number of variables of a parent solution.
Furthermore, such a variation step is only accepted when it does not degrade the fitness of the parent.
Due to the fact that variation is applied to a small number of variables, it is possible to exploit partial evaluations to greatly improve the efficiency of fitness evaluations.

GOMEA was first introduced for the domain of discrete optimization \cite{thierens2011optimal}, but has since been extended to the domains of real-valued optimization \cite{bouter2021achieving} and genetic programming \cite{virgolin2017scalable}. 

\subsection{Linkage Models}
\label{subsec:linkagemodels}
Linkage models are used by GOMEA to model the dependency structure of the optimization problem.
Such linkage models are described by a Family Of Subsets (FOS) $\mc{F} = \{\mc{F}_0, \mc{F}_1, \dots, \mc{F}_{k-q}\}$, which is a subset of the powerset of $\mc{I}$.
Therefore, it follows that $\mc{F}_i \subseteq \mc{I}$ for each $\mc{F}_i \in \mc{F}$.
Each element $\mc{F}_i$, named a linkage set, is a set containing a number of indices of problem variables that are considered to be jointly dependent.

Each variation step with the GOM variation operator considers one parent individual and one linkage set.
Variation is then applied to all variables for which the index is included in the respective linkage set.
Depending on the domain of the optimization, this variation step can consist of either crossover with a donor solution, or the sampling from a probability distribution.

\subsubsection{Marginal Products}
A Marginal Product (MP) linkage model consists of any number of non-overlapping linkage sets that together cover all problem indices 1 to $\ell$.
The simplest such linkage model is the univariate model $\mc{F}^{\texttt{Uni}} = \{\{0\},\{1\},\dots,\{\ell-1\}\}$, which models a completely separable problem.
The full linkage model $\mc{F}^{\texttt{Full}} = \{\{0,1,\dots,\ell-1\}\}$ contains one linkage set with all problem indices, and models a problem with complete dependency between each pair of variables.
Note that this linkage model can only be used in real-valued optimization, due to the nature of variation (i.e., crossover) in the discrete domain.

\subsubsection{Linkage Tree}
The Linkage Tree (LT) \cite{thierens2010linkage} is a hierarchical linkage model describing different levels of dependencies ranging from the single-variable level up to very high-level dependencies.
A linkage tree is constructed using the Unweighted Pair Group Method with Arithmetic mean (UPGMA) clustering method \cite{gronau2007optimal}.
This method is initialized with all single-variable linkage sets, and continuously merges the two (unmerged) linkage sets with the highest similarity until all linkage sets have been merged and it is left with only one linkage set containing all problem variables.
Any such merged set is added to the linkage tree, alongside all initial (univariate) linkage sets.
Therefore, the linkage tree contains all linkage sets encountered during the UPGMA process, ranging from univariate to the full linkage set.
Note that the full linkage set is removed from the linkage model in discrete optimization.

An LT may be learned during optimization based on the population, or learned \textit{a priori} based on a known similarity metric.
When it is learned during optimization, possible similarity metrics include mutual information \cite{thierens2011optimal} or hamming distance \cite{olsthoorn2021multi}.
When the LT is learned based on a known distance metric, it remains constant and is therefore named a static (or fixed) linkage tree.

\subsubsection{Conditional}
For real-valued optimization, conditional linkage models were previously introduced \cite{bouter2020leveraging}.
These linkage models allow for the application of variation to a certain subset of variables while conditioning on the values of remaining variables of the parent individual.
This mainly benefits the optimization on problems with overlapping strong dependencies.
The use of a conditional linkage model requires knowing the Variable Interaction Graph (VIG) of the optimization problem, which is only known in a GBO setting.
Furthermore, the definition of a conditional linkage model may include a Bayesian factorization to specify which variables are sampled jointly dependent, in which case it is deemed a 'Multivariate Conditional' (MCond) model.
Otherwise, the linkage model is deemed a 'Univariate Conditional' (UCond) model.

When variation is done with a specific linkage set, the variables to be sampled are conditioned on all variables connected in the VIG that are not in the respective linkage set.
Also sampling from the full probability distribution is done using the conditional factorization modeled by the VIG, using forward sampling \cite{shakya2012markov}.

As with non-conditional linkage models, the variables to which variation is applied during GOM is determined by a linkage set.
Therefore, when the FOS of the conditional linkage model has only one element consisting of all variables, GOM is only done once per generation.
Hence, this model performs 'Generational GOM' (GG).
When the FOS consists of each separate element as specified by the factorization, it is deemed to perform 'Factorized GOM' (FG).
Finally, both can be combined to what is called 'Hybrid GOM' (HG), where GG and FG are both applied every generation.
As such, a number of previously used conditional linkage models are the UCondFG, UCondGG, UCondHG, and MCondHG models \cite{bouter2020leveraging}.

\subsection{Population-sizing scheme}
By default, the \texttt{GOMEA} library uses an Interleaved Multi-start Scheme (IMS) \cite{harik1999parameter} to avoid tuning the population size, as correctly setting this can have a large impact on the performance of an EA.
Using the IMS, multiple populations of different sizes are independently subject to optimization, and run generations in an interleaved fashion.
After each $c^{\texttt{IMS}}$ generations of a population with size $n$, one generation of a population with size $2n$ is performed.
This applies to each of the interleaved populations, i.e., once the population with size $2n$ has performed $c^{\texttt{IMS}}$ generations, the population with size $4n$ will run one generation.
The initial population in IMS is initialized with the base population size $n^{\texttt{base}}$, which is generally set to the smallest reasonable population size for the respective domain.

\section{Implementation}
\label{sec:implementation}
The \texttt{GOMEA} library is mostly written in C++, with code required to interface with Python written in Cython.
Cython allows for the relatively straightforward interfacing between Python and C++.
In particular, for optimization in a GBO setting, this enables a user to write a custom GBO function in Python without a deeper understanding of the C++ optimization code.

The root package name is \texttt{gomea}, and can simply be imported into Python through \texttt{import gomea}.
This package has a number of subpackages that are structured as follows:
\begin{itemize}
\item \texttt{gomea}
\begin{itemize}
\item \texttt{discrete}
\item \texttt{real\_valued}
\item \texttt{fitness}
\item \texttt{linkage}
\item \texttt{output}
\end{itemize}
\end{itemize}

The \texttt{discrete} and \texttt{real\_valued} subpackages contain code for optimization with the discrete and real-valued variants of GOMEA, respectively.
These subpackages also each contain a \texttt{Config} class within which all input parameters for the optimization are stored, and which is passed to the C++ optimization code.

The \texttt{fitness} subpackage contains Cython classes that can be extended by a user-defined Python class for the implementation of a custom fitness function, with the following hierarchy:
\begin{itemize}
\item \texttt{FitnessFunction}
\begin{itemize}
\item GBOFitnessFunction
\begin{itemize}
\item GBOFitnessFunctionDiscrete
\item GBOFitnessFunctionRealValued
\end{itemize}
\item BBOFitnessFunction
\begin{itemize}
\item BBOFitnessFunctionDiscrete
\item BBOFitnessFunctionRealValued
\end{itemize}
\end{itemize}
\end{itemize}

Each of these Cython classes has a member variable that is a pointer to a C++ class which mirrors the Cython class.
This C++ class is instantiated during the initialization of the Cython class.
A pointer to the Cython class is passed to the constructor of the C++ class (with type \texttt{PyObject*}), and stored as a member variable.
When a (partial) evaluation is required during the optimization, a public Cython function (implemented in \texttt{EmbeddedFitness.pxi}) is called from the C++ class, and the pointer to the Cython class is passed as an argument.
Other arguments may include, e.g., a vector of problem variables and the index of a subfunction that is to be evaluated.
Within the public Cython function, the \texttt{PyObject*} pointer is typecast to one of the aforementioned Cython classes, such that its user-defined methods (overloading the default of the Cython class), e.g., the evaluation of a subfunction, can be called.

To pass the problem variables, they must be converted from a C++-type vector to a type that is interpretable by Python, for which we use the \texttt{ndarray} type included in NumPy.
This \texttt{ndarray} is initialized by creating an array wrapper around the given pointer without copying the data pointed to, which is essential to maintain the performance and scalability of GOMEA in a GBO setting, as the complexity of a partial evaluation would otherwise no longer be constant, but scale in the same way as a full evaluation.

The \texttt{linkage} subpackage includes Cython classes that can be instantiated by the user (in Python), and passed as a parameter to a \texttt{GOMEA} optimization class, indicating what type of linkage model is to be used.
Each Cython class in the \texttt{linkage} subpackage wraps a pointer to an instance of the C++ class \texttt{linkage\_config\_t}, to which all necessary parameters are passed during its instantiation.
During the construction of a \texttt{linkage\_config\_t} instance, no linkage model is yet built, but all necessary parameters are contained in this class such that a linkage model can be constructed when required by the optimization.

Finally, the \texttt{output} subpackage contains a Python class responsible for wrapping output statistics, named \texttt{OutputStatistics}, and a Cython wrapper for this class.
This Cython class wraps a pointer to an instance of a C++ class that is used to store all output statistics, and is returned by the C++ code at the end of an optimization run.
All data within this instance is copied to a member variable of the Python class \texttt{OutputStatistics}, as there is no guarantee of the lifetime of the pointer within the C++ class.
In fact, all this data is erased when the instance of the EA is used to perform another run, in which case it is undesired to lose previous output.

\section{Features}
\label{sec:features}
At the time of publication, the \texttt{GOMEA} library supports optimization with the single-objective versions of the discrete GOMEA \cite{thierens2011optimal,dushatskiy2021parameterless}, and the Real-Valued GOMEA (RV-GOMEA) \cite{bouter2021achieving}.
The (C++) source code used for these algorithms was supplied by the original authors and adapted for unification purposes and integration with the Python API.

One of the primary features of the \texttt{GOMEA} library is its compatibility with user-defined GBO functions written in Python.
This feature is elaborated on in Section \ref{subsec:customgbo}.
This section includes guidelines on how to implement such functions, and gives examples of how to implement well-known optimization functions as a GBO function.

Most linkage models, as discussed in Section \ref{subsec:linkagemodels}, previously used in literature are available in the \texttt{GOMEA} library.
This includes the filtered linkage tree \cite{bosman2013more} for discrete optimization, and conditional linkage models for real-valued optimization \cite{bouter2020leveraging}.
The use of different linkage models in the \texttt{GOMEA} library is discussed in Section \ref{subsec:implementation_linkagemodels}.
Remaining input parameters of both versions of GOMEA are specified in Section \ref{subsec:implementation_inputparams}, and the output is discussed in Section \ref{subsec:implementation_output}.

\subsection{Custom Gray-Box Optimization Function}
\label{subsec:customgbo}
The implementation of a custom GBO function according to the definition in Equation \ref{eq:gbo} requires the user to define a Python class that extends one of the following classes included in the \texttt{gomea.fitness} subpackage:
\begin{itemize}
\item \texttt{GBOFitnessFunctionDiscrete}
\item \texttt{GBOFitnessFunctionRealValued}
\end{itemize}
depending on whether the domain of the optimization problem is discrete or real-valued, respectively.
Each of these classes extends base class \texttt{FitnessFunction}, which is also present in the \texttt{gomea.fitness} subpackage.
Note that discrete optimization functions are subject to maximization and real-valued optimization functions are subject to minimization, corresponding to the conventions within these respective fields.

Such a Python class requires the user to override at least the following methods:
\mint{python}|  def number_of_subfunctions(self) -> int|
\mint{python}|  def inputs_to_subfunction(self, subfunction_index) -> np.ndarray|
\mint{python}|  def subfunction(self, subfunction_index, variables) -> float|

The method \texttt{number\_of\_subfunctions} returns the total number of subfunctions, corresponding to $q$ in Equation \ref{eq:gbo}.
The method \texttt{inputs\_to\_subfunction} returns an array indicating which variables are input for the subfunction with index \texttt{subfunction\_index}.
This corresponds to $\varmathbb{I}_{i}$ in Equation \ref{eq:gbo} when \texttt{subfunction\_index} is equal to $i$.
Finally, the method \texttt{subfunction} returns the output of the subfunction with index \texttt{subfunction\_index}, corresponding to $f_i(\vec{x}_{\varmathbb{I}_i})$ in Equation \ref{eq:gbo} when \texttt{subfunction\_index} is equal to $i$.
Note that \texttt{variables} is an array containing all problem variables $x_1$ through $x_{\ell}$ and should be indexed as such.
However, only variables contained in $\varmathbb{I}_{i}$ should be actively used for the calculation of $f_i$.
If any variable $x_u$ is used for the calculation of $f_i$ when $x_u$ is not contained in $\varmathbb{I}_{i}$, any modification of $x_u$ will not trigger the calculation of $f_i$, leading to inconsistency in the objective value.

An example of the implementation of the concatenated trap function is shown in Code Block \ref{cb:trap}.
Here, the \texttt{\_\_new\_\_} method is overridden to assign the trap size \texttt{k} as a member variable of the class, and to assert that the number of variables is a multiple of the trap size.
Furthermore, the method \texttt{inputs\_to\_subfunction} is defined such that it returns the range $[ki, \dots, k+ki]$ given the subfunction index $i$ as input, as this range defines the indices of the problem variables used by subfunction $f_i$, i.e., the $i^{\text{th}}$ trap function.
Within the method \texttt{subfunction}, these variables are then retrieved and stored into \texttt{trap\_vars}, after which they are summed using numpy.
Finally, the fitness contribution of the subfunction $f_i$ is then calculated given the calculated \texttt{unitation}, and returned.

\begin{listing}[ht]
\begin{minted}[linenos,xleftmargin=0.24in]
{python}
import gomea
import numpy as np
class ConcatTrapGBO(gomea.fitness.GBOFitnessFunctionDiscrete):
    def __new__(self, number_of_variables, k):
        assert( number_of_variables % k == 0 )
        self.k = k  # Trap size
        return super().__new__(self,number_of_variables)

    def number_of_subfunctions(self) -> int:
        return self.number_of_variables // self.k
    
    def inputs_to_subfunction(self, subf_index) -> np.ndarray:
        return range(self.k*subf_index,self.k*subf_index+self.k)

    def subfunction(self, subf_index, variables) -> float:
        trap_vars = variables[self.inputs_to_subfunction(subf_index)]
        unitation = np.sum(trap_vars)
        if unitation == self.k:
            return unitation
        else:
            return self.k - unitation - 1
\end{minted}
\caption{Concatenated trap function}
\label{cb:trap}
\end{listing}

In the concatenated trap function, the function $g$ as defined in Equation \ref{eq:gbo} is simply the identity function.
To implement an optimization function for which $g$ is not simply the identity function, this must be implemented by overriding the method:
\mint{python}|  def objective_function(self, obj_index, fitness_buffers) -> float|
This is the method corresponding to $g$ in Equation \ref{eq:gbo} which computes the fitness of an individual given an array of fitness buffer values.
By default, \texttt{fitness\_buffers} is an array containing the sum of all subfunctions at index 0.
Note that the parameter \texttt{obj\_index} is present for future compatibility with multi-objective optimization, but can remain unused in single-objective optimization.

By overriding the method \texttt{objective\_function}, it can be defined as any function of the fitness buffers of an individual.
This function does not have access to the variables of the respective solution, because this is required to be done within the implementation of \texttt{subfunction}.
Excessive access of the variables within the \texttt{objective\_function} can negate all benefits of a GBO setting and have a substantial negative impact on performance.

In order to use multiple fitness buffers, similar to the functions shown in Equation \ref{eq:gfunc_examples}, it is necessary also override the following methods, in addition to those listed at the start of Section \ref{subsec:customgbo}:
\mint{python}|  def number_of_fitness_buffers(self) -> int|
\mint{python}|  def fitness_buffer_index_for_subfunction(self, subf_index) -> int|
The method \texttt{number\_of\_fitness\_buffers} specifies the total number of fitness buffers.
The index of the fitness buffer to which the result of a subfunction (with index \texttt{subf\_index}) needs to be added is specified by \texttt{fitness\_buffer\_index\_for\_subfunction}.

The definition of a constraint function is possible in a similar way to that of an objective function, by overloading the method:
\mint{python}|  def constraint_function(self, fitness_buffers) -> float|
When this method is not overloaded, it returns 0 by default, meaning that every possible solution is feasible.

Though the most straightforward way of implementing a GBO function is in Python, it is possible to implement it in C++ for better performance.
For this, it is currently recommended to implement the custom GBO function into the class \texttt{YourFitnessFunction} (the \texttt{Discrete} or \texttt{RealValued} variant) and compiling from source.

\subsection{Custom Black-Box Optimization Function}
Similar to the definition of a GBO fitness function as discussed in Section \ref{subsec:customgbo}, also for the definition of a BBO function a class needs to be defined by the user, which in this case is required to extend 
one of the following classes in the \texttt{gomea.fitness} subpackage:
\begin{itemize}
\item \texttt{BBOFitnessFunctionDiscrete}
\item \texttt{BBOFitnessFunctionRealValued}
\end{itemize}
The choice among these classes depends on whether the domain of the optimization problem is discrete or real-valued, respectively.
Each of these classes extends base class \texttt{FitnessFunction}, which is also present in the \texttt{gomea.fitness} subpackage.

This user defined class is then only required to implement the following method, which returns the objective value of the solution defined by the input \texttt{variables}:
\mint{python}|  def objective_function(self, objective_index, variables) -> float|

Note that \texttt{objective\_index} is unused for single-objective optimization, but is required for future compatibility with multi-objective optimization.

\subsection{Linkage Models}
\label{subsec:implementation_linkagemodels}
All linkage models are implemented in the \texttt{gomea.linkage} subpackage.
The linkage models available for both real-valued and discrete optimization are the following:
\begin{itemize}
\item \texttt{Univariate()}
\item \texttt{BlockMarginalProduct(block\_size)}
\item \texttt{LinkageTree(sim\_measure,filtered,max\_set\_size)}
\item \texttt{StaticLinkageTree(max\_set\_size)}
\item \texttt{Custom(file)}
\item \texttt{Custom(fos)}
\end{itemize}
Additionally, the following linkage models are only available for real-valued optimization:
\begin{itemize}
\item \texttt{Full()}
\item \texttt{Conditional(max\_clique\_size,inc\_cliques,inc\_full)}
\begin{itemize}
\item \texttt{UCondGG()}
\item \texttt{UCondFG()}
\item \texttt{UCondHG()}
\item \texttt{MCondHG(max\_clique\_size)}
\end{itemize}
\end{itemize}
Each of these linkage models extend the base class \texttt{LinkageModel}.
The above linkage models and their parameters are discussed in this section.
The details of these linkage models are discussed in the following sections.

\subsubsection{Univariate}
This linkage model uses a FOS with $\ell$ univariate elements, and requires no input parameters.

\subsubsection{Block Marginal Product}
This linkage model uses a FOS where each element consists of \texttt{block\_size} consecutive variables starting from 0 up to $\ell-1$.
To use a marginal product FOS that does not adhere to this structure, it is advised to use the \texttt{Custom} linkage model.

\subsubsection{Linkage Tree}
The linkage tree model expects the parameters \texttt{sim\_measure} \texttt{filtered}, and \texttt{max\_set\_size}.
The parameter \texttt{sim\_measure} is a string from one of the possible options \texttt{'MI'} or \texttt{'NMI'}, indicating whether Mutual Information (MI) or Normalized Mutual Information (NMI) should be used as similarity metric.
If \texttt{filtered} is set to true, superfluous linkage sets are filtered \cite{bosman2013more}.
Finally, if \texttt{max\_set\_size} is set to any number larger than 0, no linkage sets larger than this number are formed during the linkage tree construction.

The \texttt{StaticLinkageTree(max\_set\_size)} is a linkage tree that is constant throughout optimization, and accepts only the parameter \texttt{max\_set\_size}, identical to the non-static linkage tree.
By default, this linkage tree uses connectivity within the VIG as a similarity measure.
Therefore, it cannot be used in a BBO setting.
With this linkage model, variables between which no path (of any length) exists in the VIG will never occur in the same linkage set.

A custom similarity measure can be used instead by overriding the following method of the custom fitness function that overrides a \texttt{GBOFitnessFunction} class, as discussed in Section \ref{subsec:customgbo}:
\mint{python}|  def similarity_measure(self,var_a,var_b) -> float|
This method requires as input the indices of two variables, \texttt{var\_a} and \texttt{var\_b}, and returns a similarity measure of these variables, with higher values indicating that these variables will be merged sooner in the construction process of the linkage tree.
Note that this method is expected to be symmetric, i.e., the same output is expected when the values for \texttt{var\_a} and \texttt{var\_b} are swapped.

\subsubsection{Custom}
A custom FOS requires exactly one named parameter: either \texttt{file} as a string, or \texttt{fos} as a vector of integer vectors.
The input file will have one linkage set per line, with each of its elements separated by commas or spaces.
Each element is required to be within the range $[0,\ell-1]$.

\subsubsection{Full}
The full linkage model contains one linkage set containing all problem variables, and requires no parameters.
This linkage model can only be used for real-valued optimization.

\subsubsection{Conditional}
\label{subsubsec:condlm}
Conditional linkage models are implemented in \texttt{Conditional(max\_clique\_size,inc\_cliques,inc\_full)}, where \texttt{max\_clique\_size} determines the maximum size of factors within the Bayesian factorization, as specified in Section \ref{subsec:linkagemodels}.
These factors are constructed by finding all maximal cliques up to the specified \texttt{max\_clique\_size}.
A \texttt{max\_clique\_size} equal to 1 is used by each UCond linkage model.
The boolean parameter \texttt{inc\_cliques} specifies whether each of these cliques should be included in the FOS, and the boolean parameter \texttt{inc\_full} specifies whether the full FOS element should be included.
As such, setting only the former to true means using FG, while only setting the latter to true means using GG.
Setting both to true means using HG.

\subsection{Optimization}
\label{subsec:implementation_inputparams}
The subpackages for discrete and real-valued optimization with GOMEA are named \texttt{discrete} and \texttt{real\_valued}, respectively.
An instance of either one of these algorithms can be instantiated by calling \texttt{gomea.DiscreteGOMEA} or \texttt{gomea.RealValuedGOMEA}.
These algorithms have some domain-specific input parameters that are discussed in Sections \ref{subsubsec:discrete_params} and \ref{subsubsec:realvalued_params}.
The input parameters they have in common, with potentially different default values for the Discrete (D) and Real-Valued (RV) domains, are as follows:
\begin{itemize}
\item \texttt{fitness} \hfill (required)
\begin{itemize} \item Any class with base class \texttt{FitnessFunction} from subpackage \texttt{gomea.fitness}. \end{itemize}
\item \texttt{linkage\_model} \hfill (default: \texttt{StaticLinkageTree()})
\begin{itemize} \item Any class with base class \texttt{LinkageModel} from subpackage \texttt{gomea.linkage}. \end{itemize}
\item \texttt{max\_number\_of\_populations} \hfill (default: \texttt{25})
\begin{itemize} \item Maximum number of interleaved populations within IMS. Set to 1 to disable IMS. \end{itemize}
\item \texttt{base\_population\_size} \hfill (default: \texttt{2} (D) / \texttt{10} (RV))
\begin{itemize} \item Population size of the initial population in IMS. Acts as the population size when IMS is disabled. \end{itemize}
\item \texttt{IMS\_subgeneration\_factor} \hfill (default: \texttt{4} (D) / \texttt{8} (RV))
\begin{itemize} \item Number of generations that each interleaved population performs per generation of the next largest population (with double the population size). \end{itemize}
\item \texttt{max\_number\_of\_generations} \hfill (default: \texttt{-1} (No limit))
\begin{itemize} \item Maximum number of generations that each interleaved population will perform before terminating. \end{itemize}
\item \texttt{max\_number\_of\_evaluations} \hfill (default: \texttt{-1} (No limit))
\item \texttt{max\_number\_of\_seconds} \hfill (default: \texttt{-1} (No limit))
\item \texttt{random\_seed} \hfill (default: \texttt{-1} (Randomly generated))
\end{itemize}

After instantiation of any such algorithm class (i.e., either the \texttt{DiscreteGOMEA} or \texttt{RealValuedGOMEA} class), optimization can be started by calling the \texttt{run} method, which requires no input parameters, and returns an instance of the \texttt{OutputStatistics} class.
The details of this class, containing the results of the optimization throughout each generation, are discussed in Section \ref{subsec:implementation_output}.

\subsubsection{Discrete}
\label{subsubsec:discrete_params}
The discrete GOMEA currently has no domain-specific input parameters.

\subsubsection{Real-Valued}
\label{subsubsec:realvalued_params}
The real-valued GOMEA has the following two domain-specific input parameters:
\begin{itemize}
\item \texttt{lower\_init\_range} \hfill (default: \texttt{0})
\item \texttt{upper\_init\_range} \hfill (default: \texttt{1})
\end{itemize}
These two input parameters define range between which each variable is initialized uniformly at random.

\subsection{Output}
\label{subsec:implementation_output}
The output of any run with an instance of GOMEA returns an instance of the \texttt{OutputStatistics} class, which is included in the \texttt{gomea.output} subpackage.
This class has a dictionary as member variable, named \texttt{metrics\_dict}, which is accessible using the \texttt{[]} operator, and contains lists of statistics for various metrics, where the key of the dictionary is the name of the metric, and the value is the list of data points for the metric.
When IMS is enabled, for each metric, a data point is appended at the end of every 10 generations of each (sub)population, as to not give an abundance of output.
When IMS is not enabled, for each metric, a data point is appended at the end of each generation of the EA.
As such, data points with the same index (of different metrics) correspond to the same state/point in time of the EA
By default, the following metrics are recorded, including the key used to retrieve the data from the dictionary:
\begin{itemize}
\item Number of generations \hfill (key: \texttt{generation})
\item Number of evaluations \hfill (key: \texttt{evaluations})
\item Elapsed time (seconds) \hfill (key: \texttt{time})
\item Elapsed evaluation time (seconds) \hfill (key: \texttt{eval\_time})
\item Population index \hfill (key: \texttt{population\_index})
\item Population size \hfill (key: \texttt{population\_size})
\item Best objective value \hfill (key: \texttt{best\_obj\_val})
\item Best constraint value \hfill (key: \texttt{best\_cons\_val})
\end{itemize}
The full list of metrics is accessible as the property \texttt{metrics} of the \texttt{OutputStatistics} class.

In Code block \ref{cb:plotting} example Python code is shown of how an arbitrary convergence plot can be made given the output of a run with \texttt{RealValuedGOMEA}.

\begin{listing}[ht]
\begin{minted}[linenos,xleftmargin=0.24in]
{python}
import gomea
import matplotlib.pyplot as plt
frv = gomea.fitness.RosenbrockFunction(20,value_to_reach=1e-10)
lm = gomea.linkage.Univariate()
rvgom = gomea.RealValuedGOMEA(fitness=frv, linkage_model=lm)
result = rvgom.run()
plt.plot(result['evaluations'],result['best_obj_val'])
\end{minted}
\caption{Example of plotting the output of a run.}
\label{cb:plotting}
\end{listing}

\section{Experiments}
\label{sec:experiments}
This section describes the performance and scalability of the \texttt{GOMEA} library on a number of typical benchmark problems.
Furthermore, it is shown what the benefit is of implementing an objective function in a GBO setting compared to a BBO setting.
For reproducability, code to repeat all experiments described in this section is provided in the repository of the \texttt{GOMEA} library.

\subsection{Set-up}
All experiments are performed on a server running Fedora 36 with 20 Intel(R) Xeon(R) CPU E5-2630 v4 @ 2.20GHz and 126 GB RAM.
Each run of an EA used only a single core.
Default parameters of the \texttt{GOMEA} library are used unless specified otherwise.
All plotted data points show the median and interdecile range of 30 independent successful runs.
A run is considered successful if, for discrete problems, the optimum was found, or, for real-valued problems, the value to reach of $10^{-10}$ was found.
A time limit of one hour was used, and the budget of function evaluations was set to $10^7$ for discrete problems and $10^8$ for the real-valued problem.
No experiments were performed beyond the displayed range of dimensionality.

\subsection{Benchmark functions}
We use a number of well-known benchmark functions from the domains of discrete and real-valued optimization.
This includes the concatenated deceptive trap function \cite{deb1993analyzing}, the MaxCut problem \cite{karp1972reducibility}, and the Rosenbrock function.

These benchmark problems are selected as they are from different optimization domains and/or exhibit different dependency structures.
The concatenated deceptive trap function is a well-known discrete optimization problem with strong dependencies within small disjoint subsets of variables.
We use a trap size of 5 for all experiments.
The MaxCut problem is also a discrete optimization problem, where each vertex of a given graph is assigned to a set or its complement, and the weight of edges between vertices in opposing sets is to be maximized.
We specifically use unweighted graphs with the structure of a square grid with wrap-around, i.e., a torus, leading to a non-separable optimization problem and each variable having a constant number of dependent variables.
Finally, the Rosenbrock function is a real-valued non-separable problem with overlapping dependencies.

\subsection{Results}
\begin{figure*}[h]
\centering
\begin{subfigure}{0.3\linewidth}
\includegraphics[width=\linewidth]{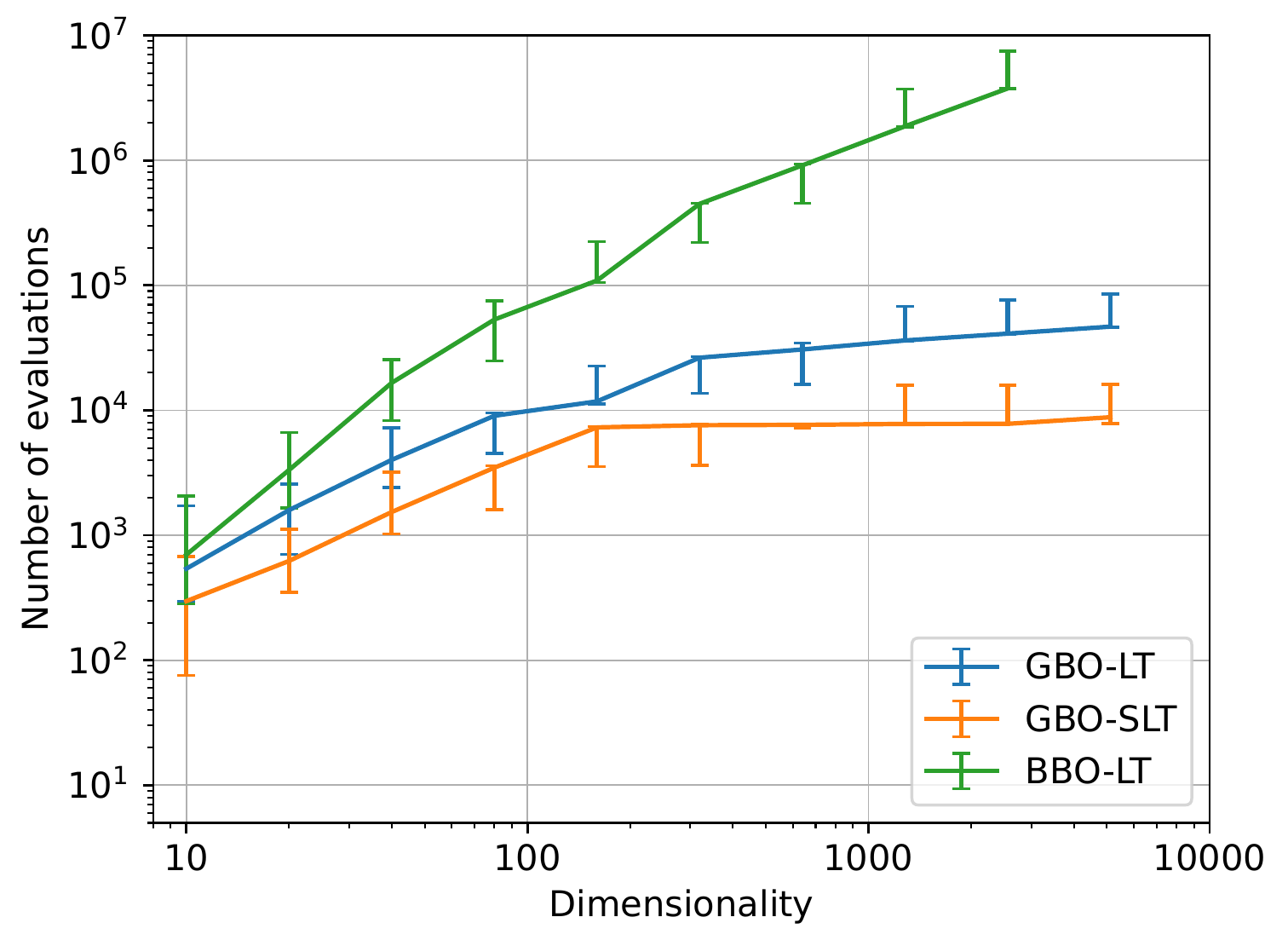}
\caption{Concatenated Deceptive Trap}
\end{subfigure}
\begin{subfigure}{0.3\linewidth}
\includegraphics[width=\linewidth]{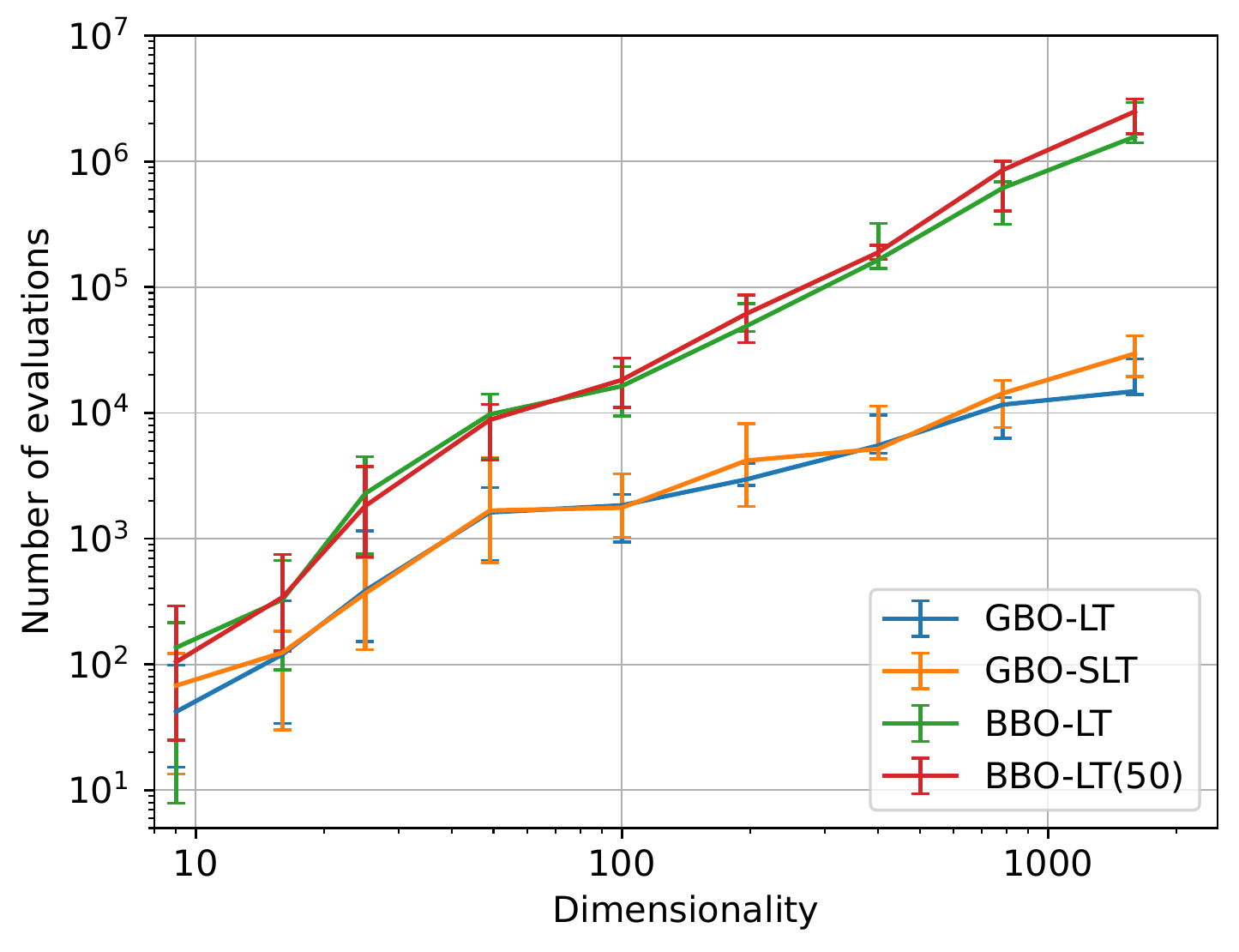}
\caption{MaxCut}
\end{subfigure}
\begin{subfigure}{0.3\linewidth}
\includegraphics[width=\linewidth]{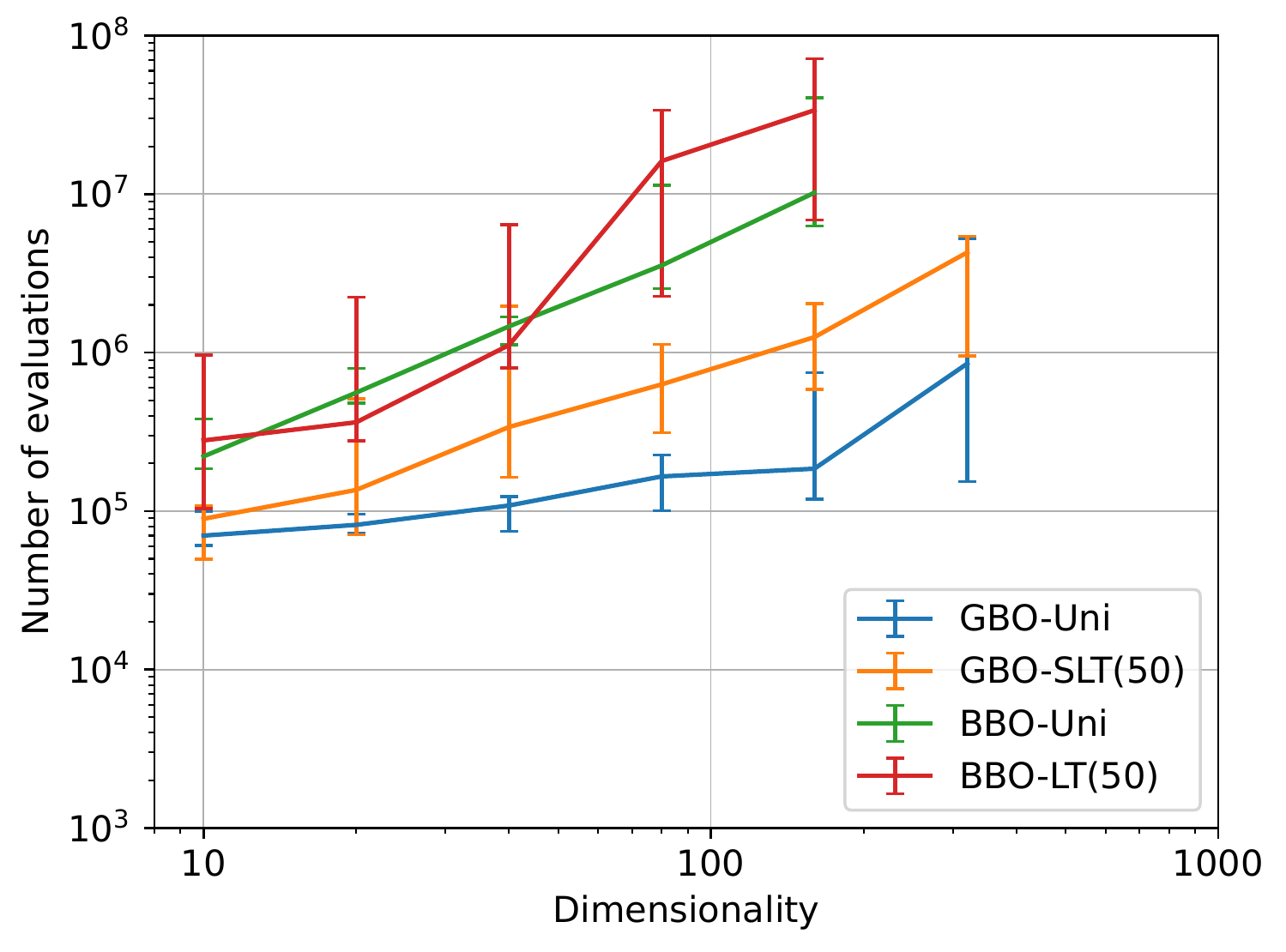}
\caption{Rosenbrock}
\end{subfigure}
\vspace{-10pt}
\caption{Scalability plots for the number of function evaluations required for different linkage models in a BBO or GBO setting. Numbers within parentheses indicate upper bound for the linkage set size.}
\vspace{-10pt}
\label{fig:scalability_fev}
\end{figure*}

Figure \ref{fig:scalability_fev} shows the scalability of the \texttt{GOMEA} library on the respective benchmark problems in terms of the number of evaluations.
Though the scalability of GOMEA on benchmark functions was previously already shown to be excellent in a GBO setting \cite{bouter2021achieving}, we here confirm these findings for the \texttt{GOMEA} library.
These results are mainly relevant for optimization problems where the overwhelming majority of computation time is spent on function evaluations.
For these kinds of problems, the reduction of the number of function evaluations by an order of magnitude would also reduce the total computation time by approximately one order of magnitude.

\begin{figure*}[h]
\centering
\begin{subfigure}{0.3\linewidth}
\includegraphics[width=\linewidth]{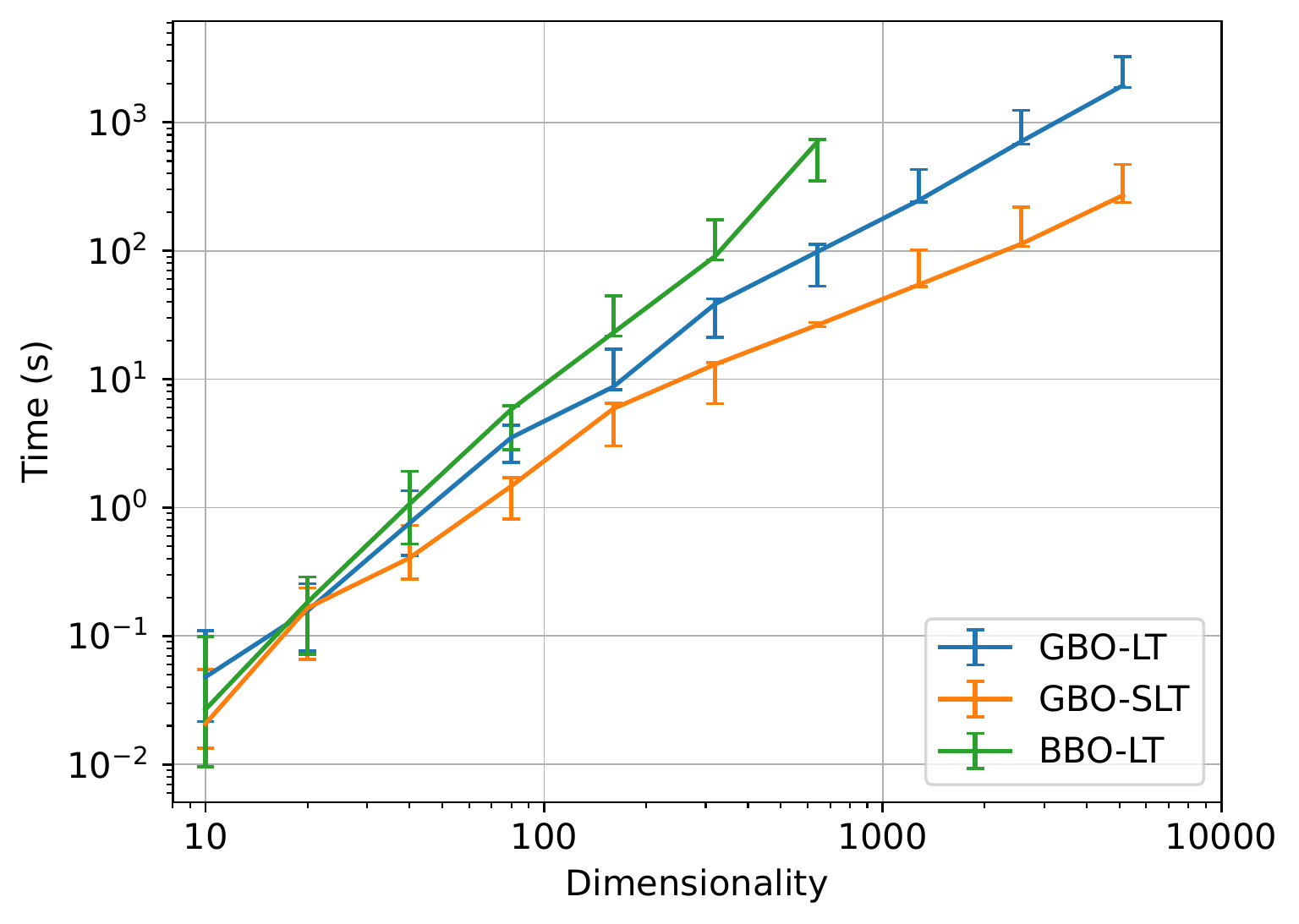}
\caption{Concatenated Deceptive Trap}
\end{subfigure}
\begin{subfigure}{0.3\linewidth}
\includegraphics[width=\linewidth]{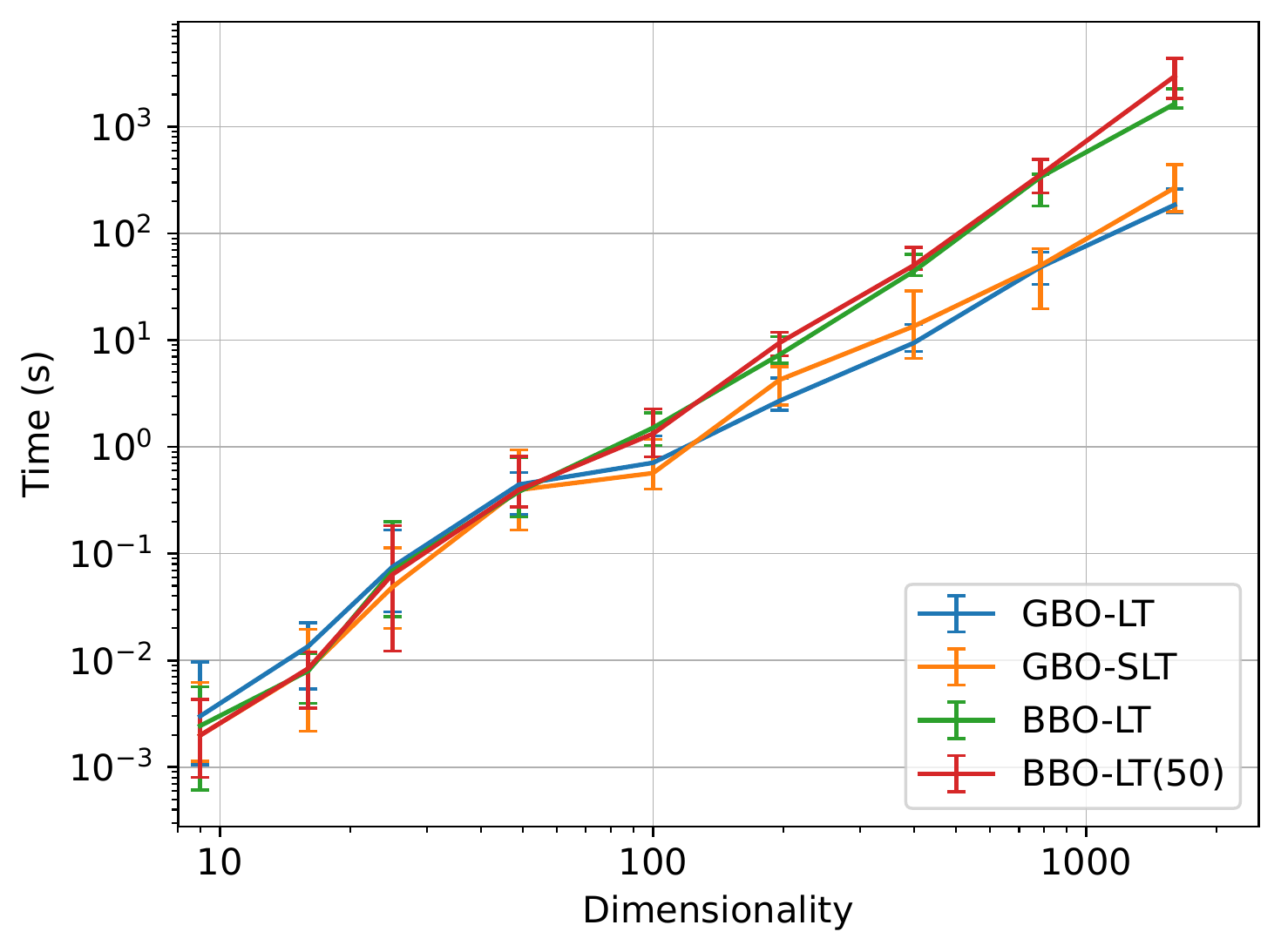}
\caption{MaxCut}
\end{subfigure}
\begin{subfigure}{0.3\linewidth}
\includegraphics[width=\linewidth]{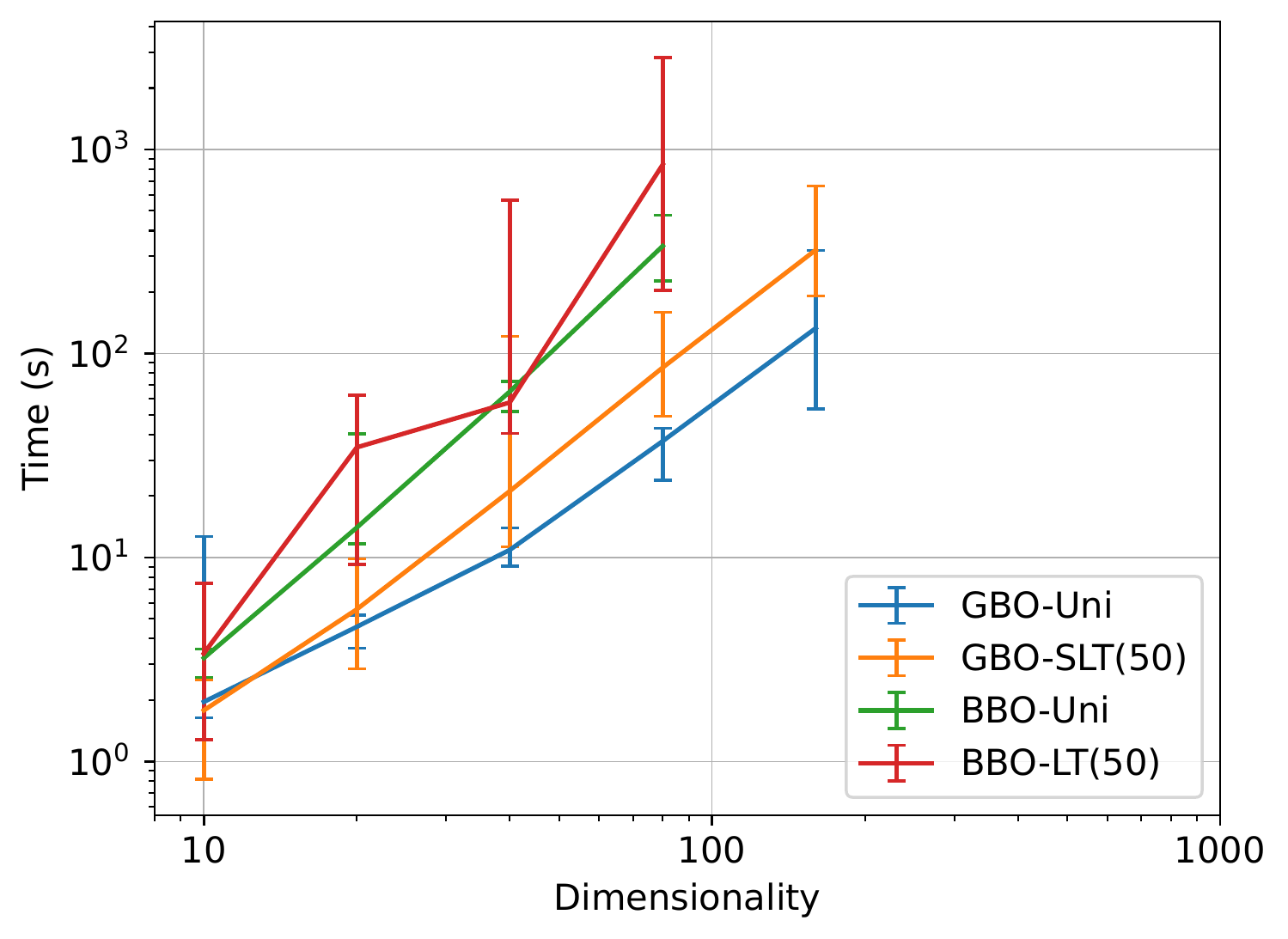}
\caption{Rosenbrock}
\end{subfigure}
\vspace{-10pt}
\caption{Scalability plots for the computation time required for benchmark functions implemented in Python for different linkage models in a BBO or GBO setting. Numbers within parentheses indicate upper bound for the linkage set size.}
\vspace{-10pt}
\label{fig:scalability_time}
\end{figure*}

In Figure \ref{fig:scalability_time}, we show the scalability of the \texttt{GOMEA} library in terms of computation time, for different linkage models in both a BBO and a GBO setting.
All problems shown are implemented in Python.
These results clearly show the benefit of using a GBO setting, mainly for large-scale optimization problems.
Moreover, the use of a GBO settings enables the use of an SLT rather than a (dynamic) LT, possibly providing additional benefit.
Finally, bounding the maximum size of linkage sets can offer a benefit, but shows no benefit on the MaxCut problem.

\begin{figure*}[h]
\centering
\begin{subfigure}{0.3\linewidth}
\includegraphics[width=\linewidth]{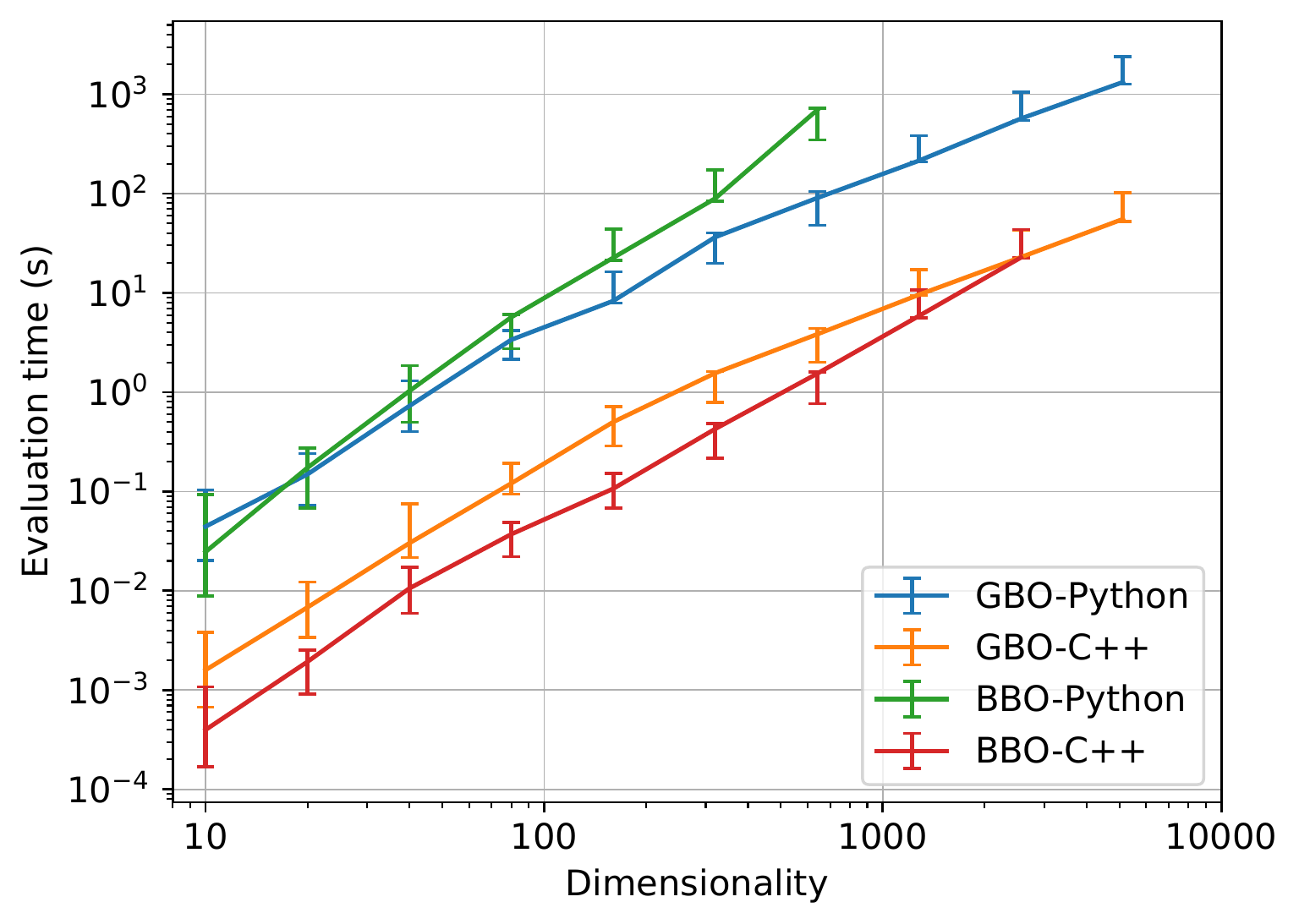}
\caption{Concatenated Deceptive Trap (LT)}
\end{subfigure}
\begin{subfigure}{0.3\linewidth}
\includegraphics[width=\linewidth]{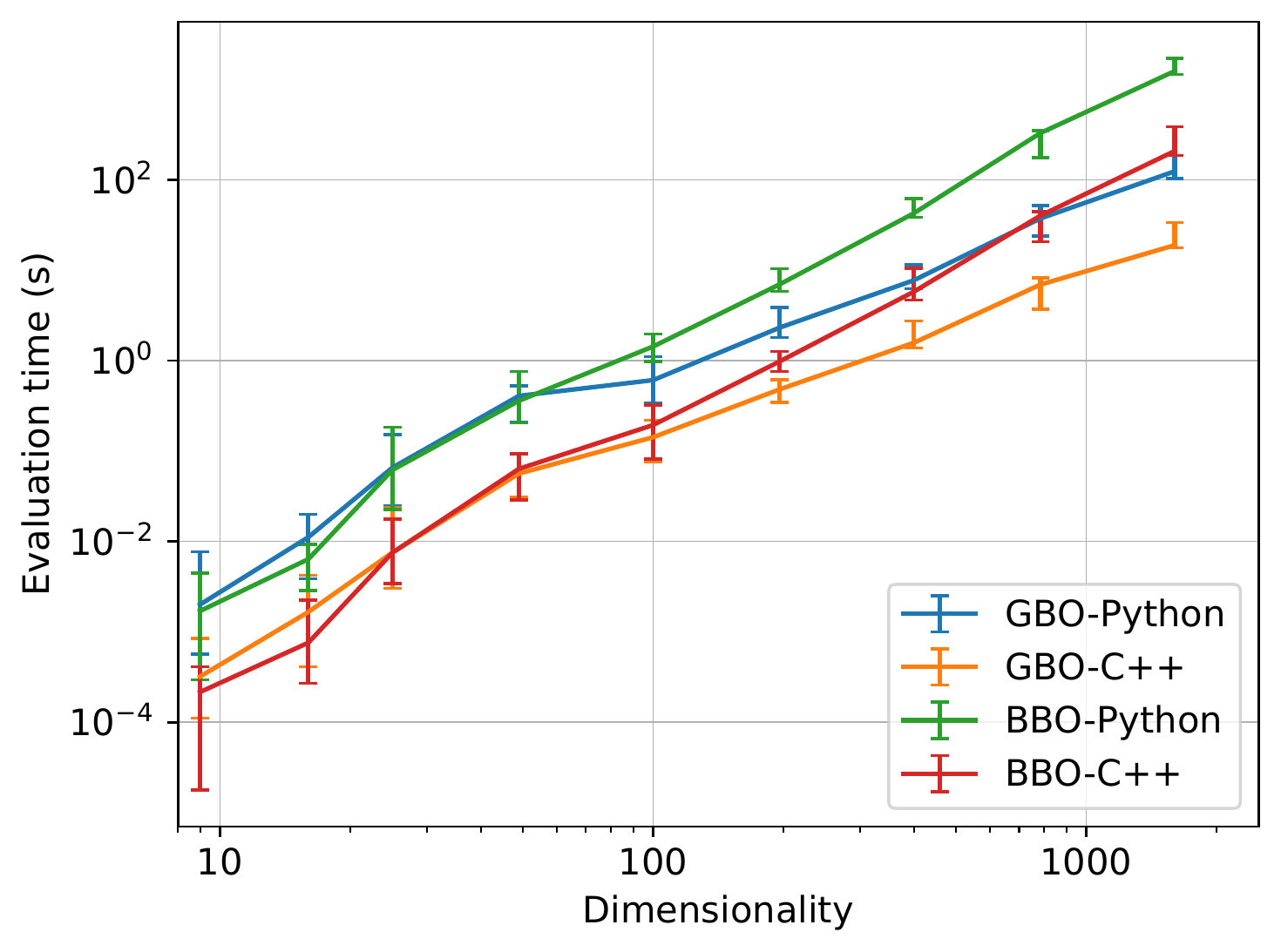}
\caption{MaxCut (LT)}
\end{subfigure}
\begin{subfigure}{0.3\linewidth}
\includegraphics[width=\linewidth]{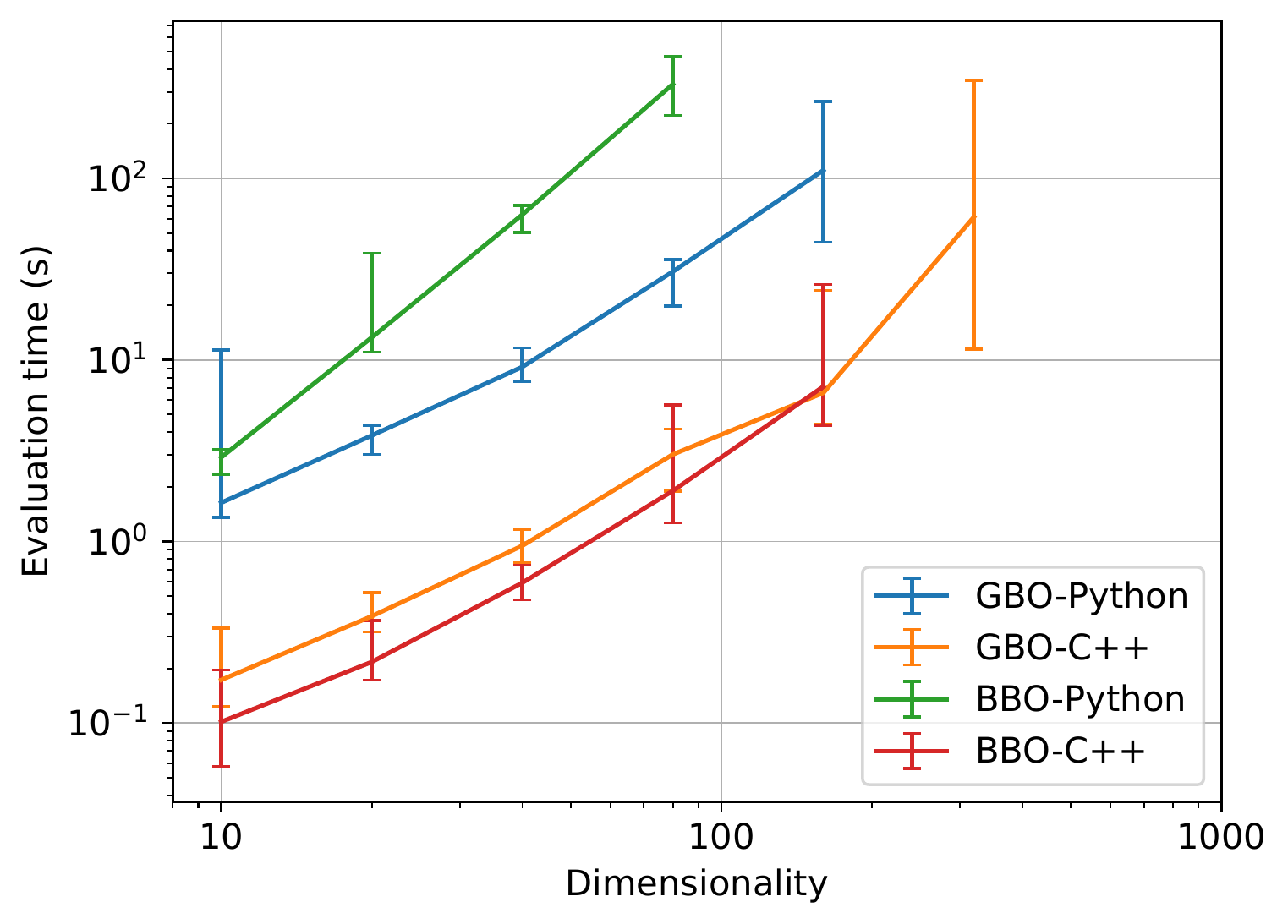}
\caption{Rosenbrock (Uni)}
\end{subfigure}
\vspace{-10pt}
\caption{Computation time spent within the evaluation function for implementations in different languages for BBO or GBO settings. Used linkage models are indicated in parentheses.}
\vspace{-10pt}
\label{fig:languages}
\end{figure*}

Though the most straightforward way of implementing an optimization function for the \texttt{GOMEA} library is using Python, it is also possible to implement such a function within the C++ code of the \texttt{GOMEA} library, as this leads to an increase in performance.
In Figure \ref{fig:languages} we show the difference in evaluation time between programming languages for BBO and GBO settings, for one linkage model per problem, to give an indication of the impact of these different implementations.
Only the time spent within the evaluation function is shown, as this is the only part of the executed code that is different.


\section{Future Work}
\label{sec:discussion}
One of the major points of future work includes the inclusion of multi-objective variants of GOMEA, for both discrete \cite{luong2014multi} and real-valued \cite{bouter2017multi} optimization, as well as the inclusing of GOMEA for the domain of genetic programming (GP) \cite{virgolin2017scalable}.
Furthermore, future work could include further customization of input parameters and output statistics of the \texttt{GOMEA} library.

\section{Conclusion}
\label{sec:conclusion}
In this paper, we introduced the \texttt{GOMEA} library, a Python library around C++ optimization code of the state-of-the-art MBEA GOMEA.
This library makes it easier for users to run GOMEA on their own user-specific problems, as the \texttt{GOMEA} library can be easily installed, and optimization functions for both BBO and GBO can be implemented in Python.
In this paper, the initial state of the \texttt{GOMEA} library was described, its structure, and how it can be used for optimization.
In our experimental results, we have shown the performance of the \texttt{GOMEA} library on various BBO and GBO benchmark problems using different linkage models.
Furthermore, the difference in performance is shown between optimization functions implemented in either Python or C++.

With the introduction of the \texttt{GOMEA} library, a large hurdle for optimization in a GBO setting has been lifted.
As such, this opens the door for users to apply GOMEA to real-world problems of their interest in a GBO setting with a much smaller time commitment.

\newpage

\bibliographystyle{ACM-Reference-Format}
\bibliography{evosoft}

\end{document}